\documentclass{article}

\usepackage{standard}
\usepackage{groups}
\usepackage{proceed2e}
\usepackage{epsfig}
\usepackage{enumerate}
\usepackage{stmaryrd}

\newcommand{\ISOtwo}{\mathrm{ISO}^{+}(2)}

\newcommand{\flm}[2]{\h f_{#1, #2}}
\newcommand{\glll}[3]{\h {\V g}_{#1, #2, #3}}
\newcommand{\Ylm}{Y_l^m}
\newcommand{\Matrix}[1]{\left( \begin{matrix} #1 \end{matrix} \right)}

\title{A novel set of rotationally and translationally invariant features for images based on the non-commutative bispectrum}
\author{
{\bf Risi Kondor}\\
\texttt{risi@cs.columbia.edu}\\
Computer Science Department, Columbia University,\\  
1214 Amsterdam Ave., New York, NY10027, USA
}

\begin{document}
\maketitle

\begin{abstract}
  We propose a new set of rotationally and translationally invariant features for image or pattern recognition and classification. 
  The new features are cubic polynomials in the pixel intensities and 
  provide a richer representation of the original image than most existing systems of invariants. 
  Our construction is based on the generalization of the concept of bispectrum to 
  the three-dimensional rotation group \m{\SO(3)}, and a projection of the image onto the sphere.
\end{abstract}

\section{Introduction}

The representation of data instances in learning algorithms is subject to the conflicting demands of 
wanting to incorprate as much information as possible about real world objects, and 
not wanting to introduce spurious information with no physical meaning. 
Image recognition is perhaps the most striking example of this phenomenon: 
clearly, the position and orientation of an object inside a larger image is purely a matter of 
representation and not a property of the object itself. 

There have been many attempts to construct rotation and translation invariant representations both in the vision community 
and in the machine learning world. 
A faithful representation of invariances is particularly important when pushing algorithms 
towards the limit of small training sets. 
When training data is abundant, it can drone out spurious degrees of freedom or average over them.  
However, in small datasets effective generalization is not possible without explicitly taking the invariances into account. 

Various types of invariants are used in signal processing and computer vison, each with its own advantages and disadvantages  
(see, e.g., \cite{book:learningwithkernels}\cite{Michaelis}). 
However, a common feature of most of these invariants is that they are lossy, in the sense 
that they do not uniquely specify the original data image. 
This becomes a particularly serious problem in discriminative learning, where the success of modern algorithms 
is to a large extent based on their ability to handle very high dimensional data, capturing as much 
information about data instances as possible. 
This is why in many cases (such as the character recognition problem to be addressed in the experimental section) 
it has often proven to be better to ignore the invariance altogether rather than risk losing valuable information 
as a side-effect of enforcing it. 

Another potential problem with existing methods is their high computational cost. 
Approaches based on summing over members of the invariance group (ghost instances, etc.) and methods 
that require an expensive kernel evaluation for each pair of instances suffer specially badly from speed issues  
(e.g., \cite{KonJeb03}). 

In this paper we propose a new class of invariant features for two dimensional images 
based on the algebra of generalized bispectra and 
a projection from the image plane onto the sphere.  
The new invariant features are strictly rotation and translation invariant (up to our bandwidth restriction and a small 
projection error), and close to complete, in the sense of 
uniquely specifying the original image up to a single rotation and translation. 
The bispectral invariants can be computed in a pre-processing step before any learning takes place 
in time \m{O(u^{5/2})}, where \m{u} is the size of the original image in pixels.  
The individual invariants are third order polynomials in the pixel intensities, and hence are relatively 
well behaved. We envisage the invariants to be used as inputs to an existing machine learning algorithm, 
for example as features to build kernels from.  
Our experiments show that using the bispectral invariants makes an immediate impact on a standard 
optical character recognition task when the training and testing intances are allowed to randomly translate and rotate. 

While the bispectrum is well known in some areas of vision and signal processing, most practicioners 
are only familiar with its classical ``Euclidean'' version \cite{Heikkila}. 
For our purposes this is not sufficient because rotations and translations together form a 
non-commutative group. 
In particular, previous work on using the bispectrum for translation and rotation invariance 
considered these two types of transformations separately, first eliminating the unknown translation and then 
the rotation from the image \cite{SadlerGiannakis}. 
While this is possible for image reconstruction, as regards generating invariant features it would amount to no more 
than transforming the image to a canonical position and orientation, which is obviosuly sensitive to variations in the image,  
since small changes can lead to vastly different optimal alignments with the canonical orientation.   

While there is a well-developed and beautiful abstract theory of bispectra on general compact groups 
developed chiefly by Ramakrishna Kakarala 
\cite{KakaralaTriple} \cite{KakaralaPhD} \cite{healy96ffts}, 
not many connections of the non-commutative case to real world problems have been explored. 
To the best of our knowledge, bispectra over non-commutative groups have never been used in the context of 
simultaneously enforcing  rotational and translational symmetries of two-dimensional images. 
The crucial new device connecting rotations and translations of the plane to the action of a compact non-commutative group 
is the projection onto the sphere proposed in this paper. 

The first half of this paper sets the scene by giving a rather abstract and general introduction to the theory of bispectra on groups. 
The second half of the paper contains our actual construction and the details of implementing it on a computer. 
The reader who is not interested in the wider context of bispectral invariants might 
find it convenient to skip directly to section \ref{sec: images}. 

\section{Bispectral Invariants}

The discrete \df{Fourier transform} of a complex-valued function \m{f\colon \cbr{\zton{n-1}} \to \CC} 
is defined 
\begin{equation}\label{eq: FT}
  \h f(k)=\sum_{x=0}^{n-1} e^{-i2\pi x k/n}\:f(x), 
\end{equation}
where \m{k} extends over \m{\zton{n-1}} and each component \m{\h f(k)} 
is the coefficient of the contribution to \m{f} at frequency 
\m{k}. 
A natural quantity of interest in signal processing is then the \df{power spectrum} 
\begin{equation}\label{eq: spectrum}
  q(k)={\h f^\ast\nts (k)}\cdot  \h f(k),
\end{equation}
where \m{{}^\ast} denotes complex conjugation. 
The power spectrum 
quantifies how much energy the signal has in each frequency band. 
Intuitively it is clear that the power spectrum should be invariant to translations of the signal. 
This is also borne out by the fact that by the convolution theorem the power spectrum is the Fourier transform 
of the \df{autocorrelation function} 
\begin{equation}\label{eq: autocorrelation}
  \mathrm{corr}(x)=\sum_{y=0}^{n-1} f(y+x)\,{f^\ast\nts(y)},
\end{equation}
(Wiener-Khinchin theorem) 
which is manifestly shift-invariant. 
Here and in the following addition and subtraction of indices and frequencies in \m{\cbr{\zton{n-1}}} 
is always to be understood modulo \m{n}. 

More formally, we define the \df{translate} of \m{f} by \m{z} as 
  \m{f^z(x)=f(x-z)}. 
Plugging into \rf{eq: FT},
\begin{multline}\label{eq: Etransl}
  \h f^z(k)=
  \sum_{x=0}^{n-1} e^{-i2\pi x k/n}\:f(x-z)=\\
  \sum_{x=0}^{n-1} e^{-i2\pi (x+z) k/n}\:f(x)=e^{-i2\pi z k/n} \h f(k), \\
\end{multline}
which shows that under translation each component of \m{\h f} is  
simply premultiplied by an \m{e^{-i2\pi z k/n}} factor.
 
The invariance of the spectrum is the result of the fact that in \rf{eq: spectrum} 
these factors cancel:
\begin{multline*}
  q^z(k)=\brbig{e^{-i2\pi z k/n}\, \h f (k)}^\ast \cdot \brbig{e^{-i2\pi z k/n}\, f(k)}=\\
  e^{i2\pi z k/n}\, {\h f^\ast\nts (k)} \,e^{-i2\pi z k/n}\, f(k)=
  {\h f^\ast\nts (k)}\cdot  f(k)=q(k).
\end{multline*}

The spectrum is often used in signal processing applications as a translation invariant characterization 
of functions. 
Unforunately, in computing the spectrum we lose all phase information:  
the spectrum only measures the energy in each band, not its phase relative to other bands. 

The idea behind bispectral invariants is to move from \rf{eq: autocorrelation} to the \df{triple correlation} 
\begin{equation*}
  a(x_1,x_2)=\sum_{y=0}^{n-1} {f^\ast\nts(y-x_1)}\: {f^\ast\nts(y-x_2)}\, f(y).
\end{equation*}
Note that in some of the literature the triple correlation is defined slightly differently, and the above quantity 
would be \m{{a^\ast(-x_1,-x_2)}}. We deviate from this convention so as to make the formulae involved in the generalization 
to groups slightly more transparent. 
Again by the convolution theorem, the (2-dimensional) Fourier transform of this function is 
\begin{equation*}
  b(k_1,k_2)={\h f^\ast\nts(k_1)}\:  {\h f^\ast\nts(k_2)}\, \h f(k_1\<+k_2), 
\end{equation*}
and this is what is called the \df{bispectrum} of \m{f}. 
Under translation \m{b} becomes 
\begin{multline*}
  b^z(k_1,k_2)=e^{i2\pi z k_1/n} {\h f^\ast\nts(k_1)} \cdot e^{i2\pi z k_2/n} {\h f^\ast\nts(k_2)}\: \cdot \\
  e^{-i2\pi z(k_1+k_2)/n}\: \h f(k_1+k_2)= b(k_1,k_2),
\end{multline*}
so the bispectrum is invariant. The remarkable fact is that unlike the ordinary power spectrum, \m{b} 
is also sufficient to reconstruct the original signal up to translation. 
The bispectrum is widely used in signal processing as a lossless shift-invariant 
representation, and various algorithms have been devised to reconstruct \m{f} from \m{b}. 

\subsection{Bispectrum on groups}\label{sec: bispgroups}

The ``Euclidean'' bispectrum introduced above would already be sufficient to construct translation invariant kernels. 
However, if we are to construct a kernel which is invariant to both translation and rotation, due to the 
intricate way in which these operations interact, we need to take a slightly more abstract viewpoint and 
re-examine what was said above from the point of view of group theory. 
While the concept of ``Euclidean'' bispectra is fairly well known in signal processing and computer vision, 
its generalization to non-commutative groups has attracted much less attention. 
The pioneering researcher in this field was  R. Kakarala \cite{KakaralaPhD}.  

Recall that a group \m{G} is a set with a multiplication operation \m{\cdot\:\colon G\times G\to G} 
obeying the following axioms:
\begin{enumerate}[G1]
\item For any \m{x,y\in G},\ \ \m{xy\tin G} \ignore{is also an element of \m{G}} (closure); 
\item For any \m{x,y,z \tin G},\ \ \m{({xy}) z=x({y z})} 
  (associativity); 
\item There is a unique element of \m{G} denoted \m{e} and called the \df{identity} 
for which \m{ex=x e=x} for any \m{x\<\in G};
\item For any \m{x\<\in G} there is a corresponding element \m{x^{-1}\<\in G}  
  called the \df{inverse} of \m{x}, which satisfies \m{x x^{-1}=x^{-1} x=e} for any \m{x\in G}. 
\end{enumerate}
Significantly, groups need not be commutative, i.e., \m{x y} need not equal \m{y x}. 
This is crucial for our present purposes since rigid planar motions don't commute. 

Given a group \m{G} and a function \m{f\colon G\to \CC} to define the Fourier transform of \m{f}  
we need to introduce the concept of \df{group representations}. 
A representation is essentially a way of modeling the group operation by the multiplication of 
complex valued matrices. 
We say that \m{\rho\colon G\to\CC^{d_\rho\times d_\rho}} is a representation of \m{G} if 
\begin{equation*}
  \rho(xy)=\rho(x)\ts\rho(y) 
\end{equation*}
for any \m{x,y\in G}. We also require \m{\rho(e)\<=I}.
We say that \m{d_\rho} is the {dimensionality} of the representation. 
Note that 
\m{\rho(x^{-1})=\br{\rho(x)}^{-1}}. 
    
There are some trivial ways of producing new representations from existing ones. 
For example, if \m{\rho_1} is a representation of \m{G}, then 
for any invertible matrix \m{T},  so is \m{T^{-1}\nts \rho_1(x)\, T}. 
These representations are clearly not substantially different, so they are called \df{equivalent}. 

Another way that representations may be related 
is when a larger representation splits into smaller ones. 
We say that \m{\rho} is \dfsub{reducible}{representation} 
if some invertible square matrix \m{T} can block diagonalize it in the form 
\begin{equation*}
  T^{-1}\nts \rho(x)\, T=
  \br{\begin{array}{c|c}\rho_1(x)&0\\ \hline 0 & \rho_2(x)\end{array}} \qqquad x\in G
\end{equation*}
into a direct sum of smaller representations \m{\rho_1} and \m{\rho_2}. 

To develop the theory what are really important are the \df{irreducible} representations 
that cannot be reduced in this way. 
Given a group \m{G} there is a lot of interest in constructing a complete set 
of inequivalent representations for it. Such a set we will denote by \m{\Rcal}. 
For a wide range of groups we can choose \m{\Rcal} to consist exclusively of unitary 
representations, so from now on we assume that \m{\rho(x^{-1})=\rho(x)^\dag}, 
where \m{{}^\dag} denotes the conjugate transpose.  

With these concepts of representation theory in hand, we return to \rf{eq: FT} and note that 
the exponential factors appearing in the summation are nothing but representations 
(specifically, one-dimensional, irreducible representations) of the group formed by \m{\cbr{\zton{n-1}}} 
with respect to addition modulo \m{n}. 
This suggests generalizing Fourier transformation to the non-commutative realm in the form 
\begin{equation}\label{eq: FT gen} 
\h f(\rho)=\sum_{x\in G} f(x)\Ts  \rho(x)\qqquad \rho\in\Rcal. 
\end{equation} 
Here and in the following the summation sign either denotes a discrete sum over the elements of a discrete group, 
or an integral (with respect to Haar measure) over a Lie group.  
Note that in contrast to \rf{eq: FT}, for general groups the components of \m{\h f} are 
matrices and not scalars, and they are not 
indexed by the elments of \m{G}, but by its irreducible representations. 

The generalized Fourier transform shares many important properties with its Euclidean counterpart, 
but most of these will not concern us here. 
What is important is that there is a natural concept of translation of functions on \m{G} defined by 
\begin{equation*}
  f^z(x)=f(z^{-1} x)\qqquad z\in G,
\end{equation*}
and that by the defining property of representations,
\begin{multline*}
  \h f^z(\rho)=\sum_{x\in G} f(z^{-1}x)\Ts  \rho(x)=\\
  \sum_{x\in G} f(z^{-1}x)\Ts  \rho(z) \rho(z^{-1} x)=\\
  \rho(z)\sum_{x\in G} f(x)\Ts  \rho(x)=
  \rho(z)\h f(\rho)
\end{multline*}
in exact analogy with \rf{eq: Etransl}. 
In particular, 
by the unitarity of \m{\rho}, 
the generalized power spectrum 
  \m{q(\rho)=\h f(\rho)^\dag \h f(\rho)}  
is again invariant to translation:
\begin{multline*}
  q^z(\rho)=\brbig{\rho(z)\h f(\rho)}^\dag \brbig{\rho(z)\h f(\rho)}=\\
   f(\rho)^\dag \rho(z)^\dag \rho(z) f(\rho)^\dag =\h f(\rho)^\dag \h f(\rho).
\end{multline*}
As in the classical case, the power spectrum does not uniquely determine \m{f}. 
The loss of information is related to the fact that the 
\m{q(\rho)} matrices are by definition constrained to be positive definite,  
and again the power spectrum is insensitive to phase information in the sense that we may multiply any 
Fourier component by a different invertible matrix without affecting the power spectrum. 

To construct the bispectrum we need to couple the different components of \m{\h f}, while at the same time 
retaining invariance. Consider tensor products \m{\h f(\rho_1)\otimes \h f(\rho_2)}, which transform 
according to 
\begin{equation*}
  \h f^z(\rho_1)\otimes \h f^z(\rho_2)=\br{\rho_1(z)\otimes \rho_2(z)} \brbig{\h f(\rho_1)\otimes \h f(\rho_2)}. 
\end{equation*}
Now \m{\rho_1(z)\otimes \rho_2(z)} is also a representation of \m{G}, but typically it is not irreducible. 
However, for wide classes of groups tensor product representations decompose into irreducibles in 
the form 
\begin{equation*}
  \rho_1(z)\otimes \rho_2(z) = C\: \sqbBig{\bigoplus_{\rho} \rho(z) }\:C^\dag. 
\end{equation*}
Determining which set of irreducibles the direct sum ranges over (and with what multiplicities) and 
what the unitary matrix \m{C} should be is in general a highly non-trivial problem in representation theory. 
For now we assume that this so-called Clebsch-Gordan decomposition is known.    

In this case we have a generalized bispectrum 
\begin{equation}\label{eq: gen bispectrum}
  b(\rho_1,\rho_2)=C^\dag\: \brbig{\h f(\rho_1)\otimes \h f(\rho_2)}^\dag\:  C\: \bigoplus_{\rho}\: \h f(\rho),  
\end{equation}
and it will be translation invariant, \m{b^z(\rho_1,\rho_2)=b(\rho_1,\rho_2)}. 
What goes beyond a straightforward generalization of the classical results is the proof that 
for a wide range of groups, including all compact groups, if all \m{\h f(\rho)} Fourier components are invertible matrices, then \m{b} 
uniquely determines \m{f} up to translation. This is a highly technical result proved in \cite{KakaralaPhD}, and in constrast to 
the commutative case, there might not be an algorithm for recovering \m{f}. 

\subsection{Homogeneous spaces}

Before addressing the problem of image invariants, we need one more technical extension of the foregoing. 
We say that a group \m{G} \df{acts} on a space \m{X}, if for any \m{g\<\in G} there is a mapping \m{T_g\colon X\to X} 
such that if \m{g_2 g_1=g_3}, then \m{T_{g_1}(T_{g_2}(x))=T_{g_3}(x)} for any \m{x\in X}. 
Now \m{X} is a \df{homogeneous space} of \m{G} if fixing any \m{x_{0}\in X}, the set \m{T_g(x_0)} ranges over the 
whole of \m{X} as \m{g} ranges over \m{G}. The classical example of a homogenous space,   
which will also be our choice for our image recognition problem, 
is the unit sphere \m{S_2}. The sphere is a homogeneous space of the three-dimensional rotation group 
\m{\SO(3)}: taking the North pole as \m{x_0}, a suitable rotation can move it to any point \m{x\in S_2}. 

Fourier transformation generalizes naturally to functions \m{f\colon X\to \CC}:
\begin{equation*}\label{eq: FT homo} 
\h f(\rho)=\sum_{g\in G} f(T_g(x_0))\Ts  \rho(g)\qqquad \rho\in\Rcal,
\end{equation*} 
as does the concept of translation, \m{f^g(x)=f(T_{g^{-1}}(x))}, 
and the bispectrum \rf{eq: gen bispectrum} remains invariant to such translations. 

Note that except for the trivial case \m{X=G}, Fourier transforms on homogeneous spaces are 
naturally redundant: typically \m{X} is a much smaller space than \m{G}, yet a Fourier transform on \m{X} has the same number of 
components as a Fourier transform on the entire group. 
One manifestation of this fact is that we might find that some Fourier components are rank deficient 
no matter what \m{f\colon X\to\CC} we choose. 
While this destroys Kakarala's uniqueness result, in practice we often find that the bispectrum still 
furnishes a remarkably rich invariant representation of \m{f}. 
We remark that that invariance to right-translation \m{f^{(z)}(x)=f(x z^{-1})} would be a different matter: 
there is a variant of the bispectrum which retains the uniqueness property in this case  
(Theorem 3.3.6 in \cite{KakaralaPhD}).  
 


\section{Bispectral invariants for images}\label{sec: images}

After the abstract discussion of the previous section we now set out to construct concrete invariants 
for 2D monochrome images. 
We represent an image as an intensity function \m{h\colon \RR^2\to\sqb{0,1}} 
with support confined to a compact region of the plane, 
for example, the square \m{\sqb{-0.5,0.5}^2}. 
The group that we would ideally like to be working with encompassing all translations and rotations 
is the Euclidean group \m{\mathrm{ISO}^{+}\!\tts(2)} of rigid body motions in the plane. 
\m{\RR^2} is a homogeneous space of \m{\mathrm{ISO}^{+}\!\tts(2)}, so we could compute the \m{\mathrm{ISO}^{+}\!\tts(2)}-Fourier 
transform of our image, and construct its bispectrum as described above. 

The problem with this approach is that \m{\mathrm{ISO}^{+}\!\tts(2)} is not compact. 
Although it does belong to a class of exceptional groups to which Kakarala's 
uniqueness result does apply, its representation theory is complicated and computing the bispectrum 
is likely to be computationally very challenging.  
The main contribution of this paper is to show how the reduce the problem to 
rotations of the sphere. 
The rotation group \m{\SO(3)} also happens to have the simplest and best known non-trivial   
Clebsch-Gordan decomposition. 
To make the exposition as elementary as possible, we derive the bispectral invariants from 
first principles, exploiting the simplifications afforded by this special case.    

\subsection{Projection onto the sphere}

We begin by projecting our image \m{h} onto the unit sphere \m{S_2}. 
The simplest possible projection is to project parallel to the \m{z}-axis, formally
\begin{equation}\label{eq: mapping}
  h\mapsto f, \qquad 
  f(\theta,\phi)=h(r_{\RR^2},\theta_{\RR^2})=h(\tovr{a} \theta,\phi),  
\end{equation}
where \m{ 0\<\leq \theta \< \leq \pi} and \m{0\<\leq \phi\< < 2\pi} are spherical polar coordinates, 
while \m{r_{\RR^2}\<=\ovr{a}\theta} and \m{\theta_{\RR^2}\<=\phi} are planar polars. 
The magnification parameter \m{a} we are free to choose between reasonable bounds as long as our image ``fits'' 
on the surface of the sphere. 
Inevitably, such a mapping does involve some distortion, particulary at the corners, 
as the image conforms to the curved surface of \m{S_2}. 
Reducing \m{a} decreases this distortion at the expense of reducing the surface area of the sphere 
actually occupied by the image, and hence increasing the computaional cost at 
the same effective resolution.   
In practice, even relatively large values of \m{a} (up to \m{1.5}) do not hurt performance. 
Apart from the inevitable finite bandwidth cutoff, this is the 
only approximation involved in our method. 

\begin{figure}[tbp]
  \centering
  \includegraphics[width=.4\textwidth]{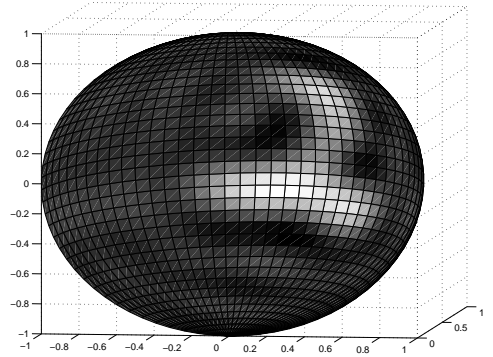}
  \caption{A NIST handwritten digit projected onto the sphere. The band-limit is \m{L=15}. 
  Note that there is a minimal amount of ``ringing''.}
  \label{fig: projected}
\end{figure}

To numerically represent \m{f} we use \df{spherical harmonics} 
\begin{equation*}
  Y_l^m(\theta,\phi)= \sqrt{\fr{2l+1}{4\pi}\fr{\br{l-m}!}{\br{l+m}!}}~~ P_l^m(\cos \theta)~ e^{im\phi},  
\end{equation*}
where \m{l=0,1,2,\ldots};~~\m{m=-l,-l+1,\ldots,l} and \m{P_l^m} are the associated Legendre polynomials.  
Recall that the spherical harmonics are the eignefunctions of the Laplace operator on \m{S_2} (with eigenvalue \m{-l^2}), 
and they form an orthonormal basis for \m{L_2(S_2)}, 
thus we can represent \m{f} as 
\begin{equation}\label{eq: Ylm exp}
  f(\theta,\phi)=\sum_{l=0}^{\infty} \sum_{m=-l}^{l} \flm{l}{m}\, Y_l^m(\theta,\phi)
  \qqquad 
\end{equation}
where \m{\flm{l}{m}=\inp{f,Y_l^m}} and \m{\inp{\cdot,\cdot}} is the inner product 
\begin{equation*}
  \inp{f,g}=\int_{0}^{\pi} \int_0^{2\pi} {f^\ast\nts(\theta,\phi)}\, g(\theta,\phi) \cos\theta \:d\phi\: d\theta. 
\end{equation*}
We denote by \m{\V {\h f}_l} the vector \m{\brN{\flm{l}{-l},\flm{l}{-l+1},\ldots,\flm{l}{l}}}. 

Viewing \m{S_2} as a homogeneous space of \m{\SO(3)}, 
the \m{\cbrN{\flm{l}{m}}} are the {Fourier coefficients} of \m{f\colon S_2\to\CC} 
as defined in the previous section.   
However, in this special case they do not form matrices, only vectors:  
if we formally computed \rf{eq: FT homo}, we would find that only the first column of each matrix is non-zero 
(see also \cite{healy96ffts}). 
This will make the computational burden significantly lighter. 

In a computational setting we must truncate \rf{eq: Ylm exp} at some finite \m{L}, 
preferably so as to match the resolution of our original image. 
In general, the spherical representation of an image requires more storage than the original 
pixmap representation only to the extent that the image only occupies a fraction of the surface of the sphere. 


For a \m{\sqb{0,1}}-valued bitmap matrix M, the mapping \rf{eq: mapping} leads to  
\begin{equation}\label{eq: mapping exp}
  \flm{l}{m}=\sum_{i,j=1}^{n} M_{i,j} \Ylm(\theta,\phi), 
\end{equation}
where \m{\theta=a \sqrt{x^2+y^2}}, 
\begin{equation*}
  \phi=
  \begin{cases}
    ~\arctan(y/x) & \text{if~~} y\<>0\\ 
    ~2\pi-\arctan(y/x) & \text{if~~} y\<<0\\ 
  \end{cases},
\end{equation*}
and \m{\br{x,y}=\brbig{\,\fr{i-1/2}{N}-0.5,~ \fr{j-1/2}{N}-0.5}}. 

Just as the isometry group of \m{\RR^2} is \m{\ISOtwo}, the isometry group of \m{S_2} is \m{\SO(3)}, 
the group of rotations of \m{\RR^3} about the origin. It is easy to visualize that given the mapping \rf{eq: mapping},
locally, around the north pole, 
there is a one-to one corresponence between the action of \m{\SO(3)} on functions on the sphere and of 
\m{\ISOtwo} on the corresponding functions on the plane. 
In other words, any rigid motion of an image in the plane can be imitated by a 3D rotation of the corresponding 
function on \m{S_2}.   
Rotations of the image around the center of the image correspond to rotations of the sphere about 
the \m{z} axis (pole to pole), while translations correspond to rotations around the \m{x} and \m{y} axes. 
Exploiting this fact, we proceed by computing the bispectral invariants of \m{f} with respect to \m{\SO(3)} 
and let these be our translation and rotation invariant features.

\subsection{An \m{\SO(3)}-invariant kernel on \m{L_2(S_2)}}

To construct the \m{\SO(3)}-invariant features, we examine how \m{\SO(3)} acts on individual spherical harmonics. 
Since \m{\cbr{\Ylm}_{m=-l,\ldots,l}} span the space of eigenvectors of the Laplace operator with eigenvalue \m{-l^2}, and 
since the Laplace operator is rotationally invariant, 
under the action of a rotation 
\m{R\in \SO(3)},~ \m{\Ylm} must transform into a linear combination \m{R(\Ylm)=\sum_{m'=-l}^{l} a_m Y_l^{m'}} of other 
spherical harmonics of the same order \m{l}. 

For a general function \m{f\in L_2(S_2)}, under a rotation \m{R\in\SO(3)} the Fourier coefficients transform according to 
\begin{equation}\label{eq: trafo1}
  \Matrix{\flm{l}{-l}'\\ \vdots \\\flm{l}{l}'} = D^{(l)}(R)  \Matrix{\flm{l}{-l}\\ \vdots \\\flm{l}{l}},  
\end{equation}
where \m{D^{(l)}(R)} are  \m{(2l\<+1)\times(2l+1)} dimensional matrices. 
In fact, \m{D^{(0)},D^{(1)},\ldots } are exactly the (complex-valued) irreducible representations of \m{\SO(3)}.    

\ignore{
Recall that for any group \m{G}, we say that a matrix-valued function \m{D} is a \df{representation} of \m{G} if 
for any \m{g_1,g_2\in G},~\m{D(g_2) D(g_1)\<=D(g_2 g_1) }. The representation is called \df{irreducible} if 
there is no invertible square matrix \m{T} reducing \m{D} to block-diagonal form  
\begin{equation*}
      T^{-1}\nts D (g)\, T=
  \br{\begin{array}{c|c}D_1(g)&0\\ \hline 0 & D_2(g)\end{array}} \qquad \forall\;g\in G. 
\end{equation*}
}

\begin{figure}[tbp]
  \centering
  \includegraphics[width=0.3\textwidth]{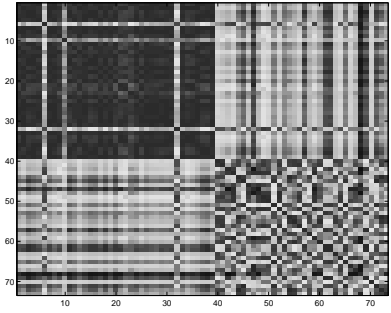}
  \caption{The inner product matrix between the bispectrum representation of the "0" and "1" digits from the first 300 
    translated and rotated NIST characters. 
  The block structure reflects that the intra-class inner products are higher than the inter-class products.} 
  \label{fig: Gram}
\end{figure}

It is possible to show that the \m{D^{(l)}} are unitary representations, hence 
the polynomials   
\begin{equation*}
  p_{l}=\sum_{m=-l}^{l} \absbig{\flm{l}{m}}^2= \V{\h f}_l^\dagger \cdot \V{\h f}_l = 
  \Matrix{\flm{l}{-l}^\ast,  \hdots, \flm{l}{l}^\ast}\cdot  \Matrix{\flm{l}{-l}\\ \vdots \\\flm{l}{l}}
\end{equation*}
transform according to 
\begin{multline*}\label{eq: power spectrum}
  p_l\mapsto \brbig{D^{(l)}(R)\, \V{\h f}_l}^\dag \cdot \brbig{D^{(l)}(R)\, \V{\h f}_l}=\\
  \V{\h f}_l ^\dag \,\brN{D^{(l)}(R)}^\dagger\, \brN{D^{(l)}(R)}\, \V{\h f}_l = \V{\h f}_l^\dagger \cdot \V{\h f}_l,
\end{multline*}
i.e., they are invariant. 
This is the power spectrum, as defined in Section \ref{sec: bispgroups}. 
As before, this is an invariant, but very impoverished representation of \m{f}. 

\begin{figure*}[t!]
  \centering
  \includegraphics[width=.1\textwidth]{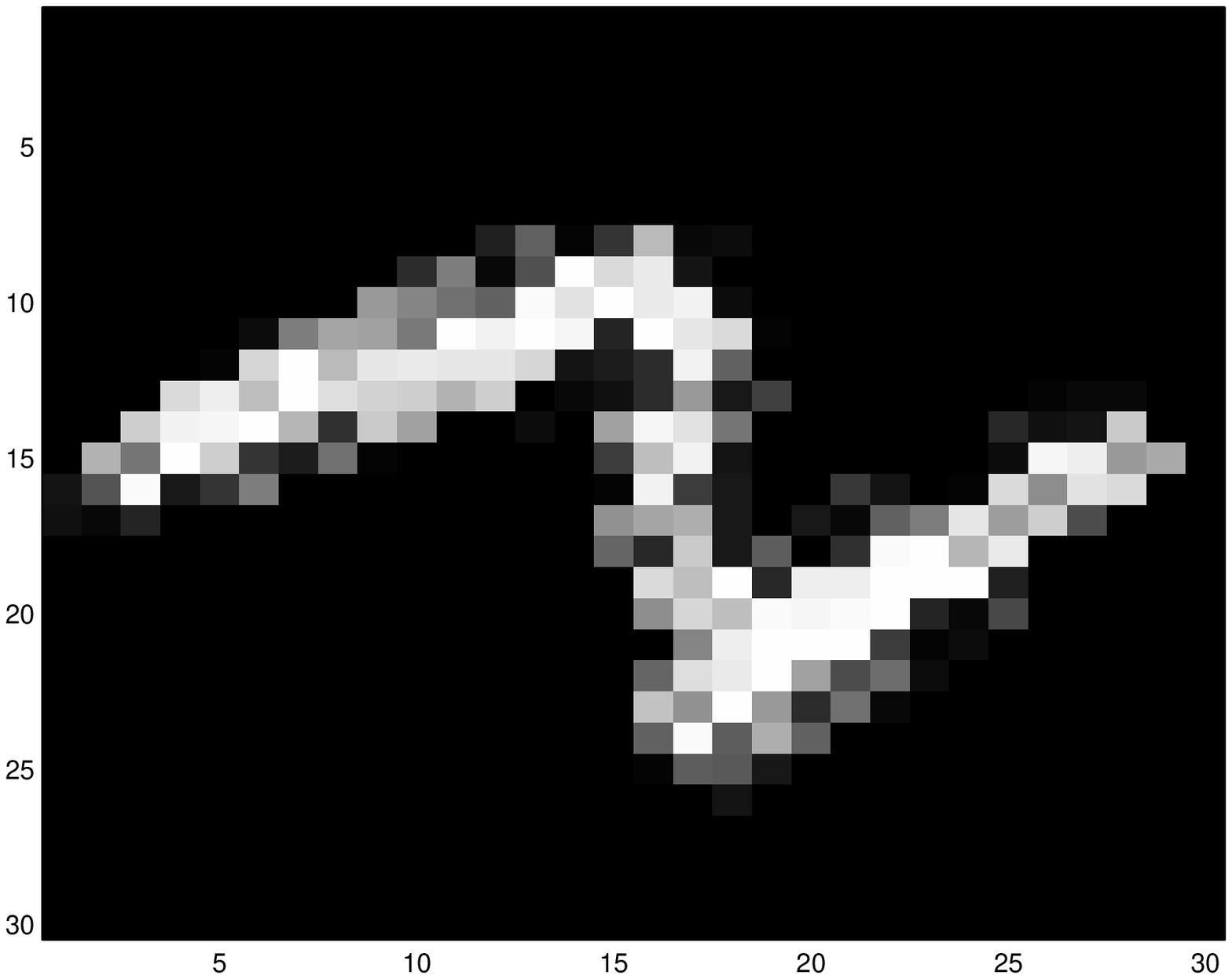}
  \includegraphics[width=.1\textwidth]{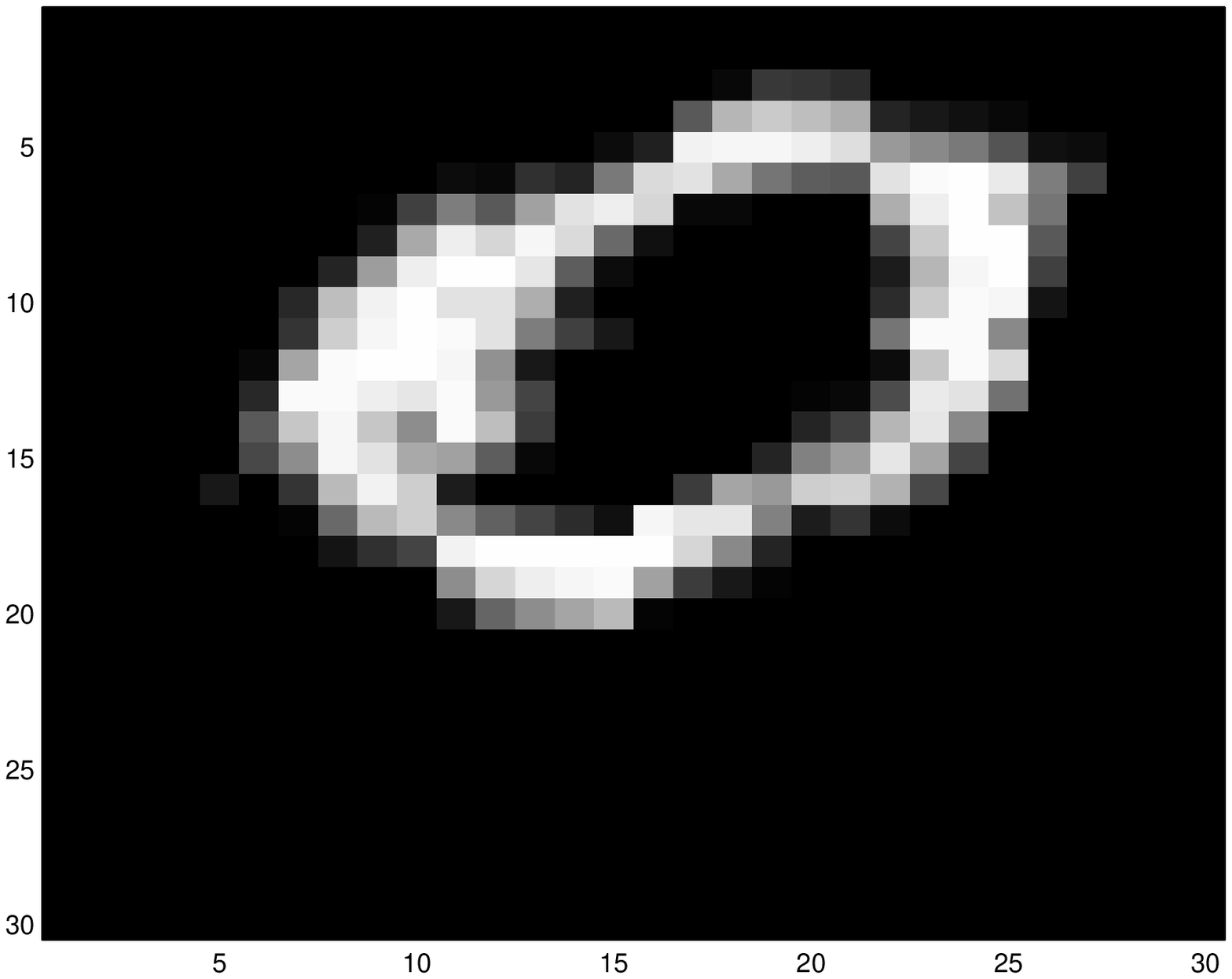}
  \includegraphics[width=.1\textwidth]{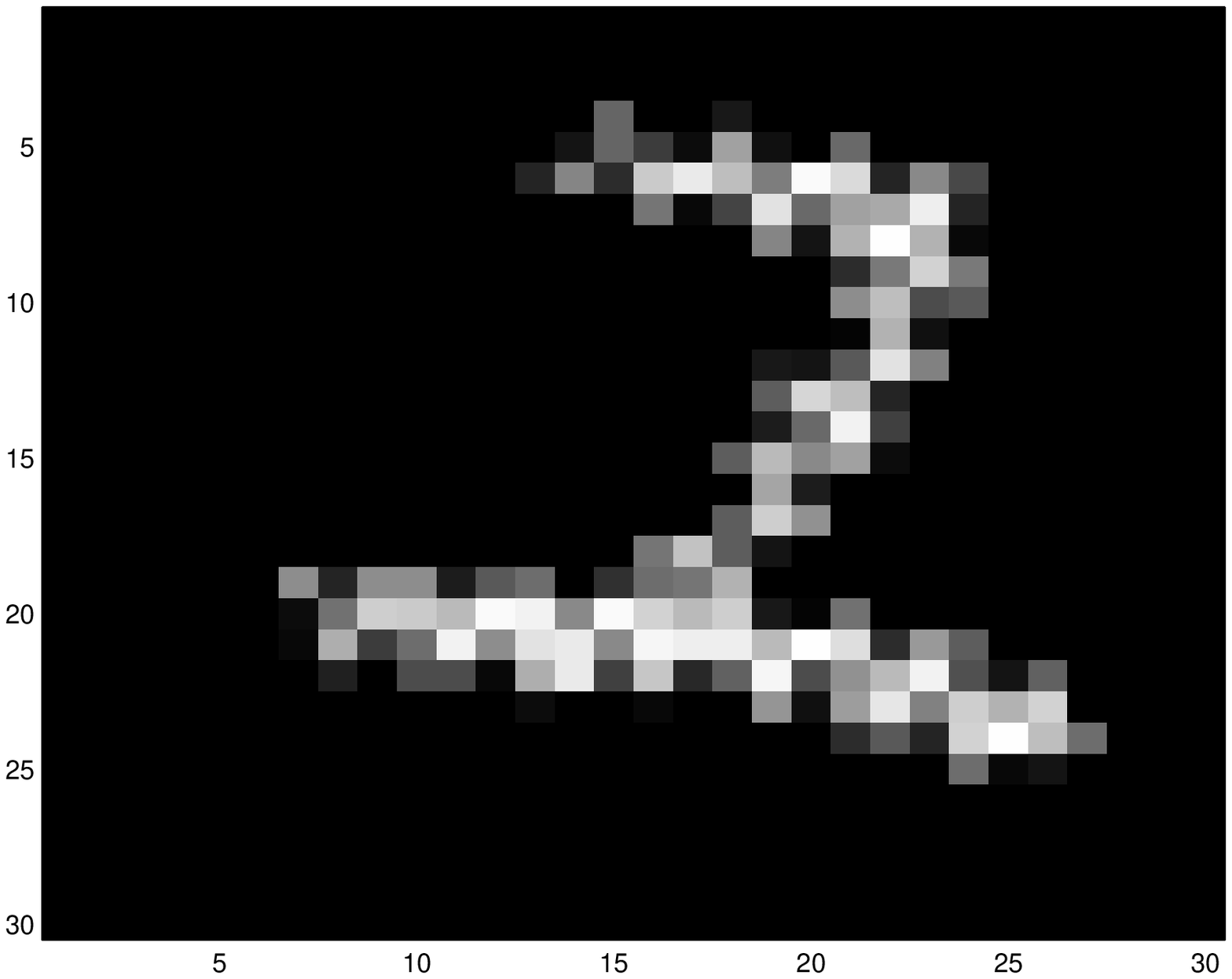}
  \includegraphics[width=.1\textwidth]{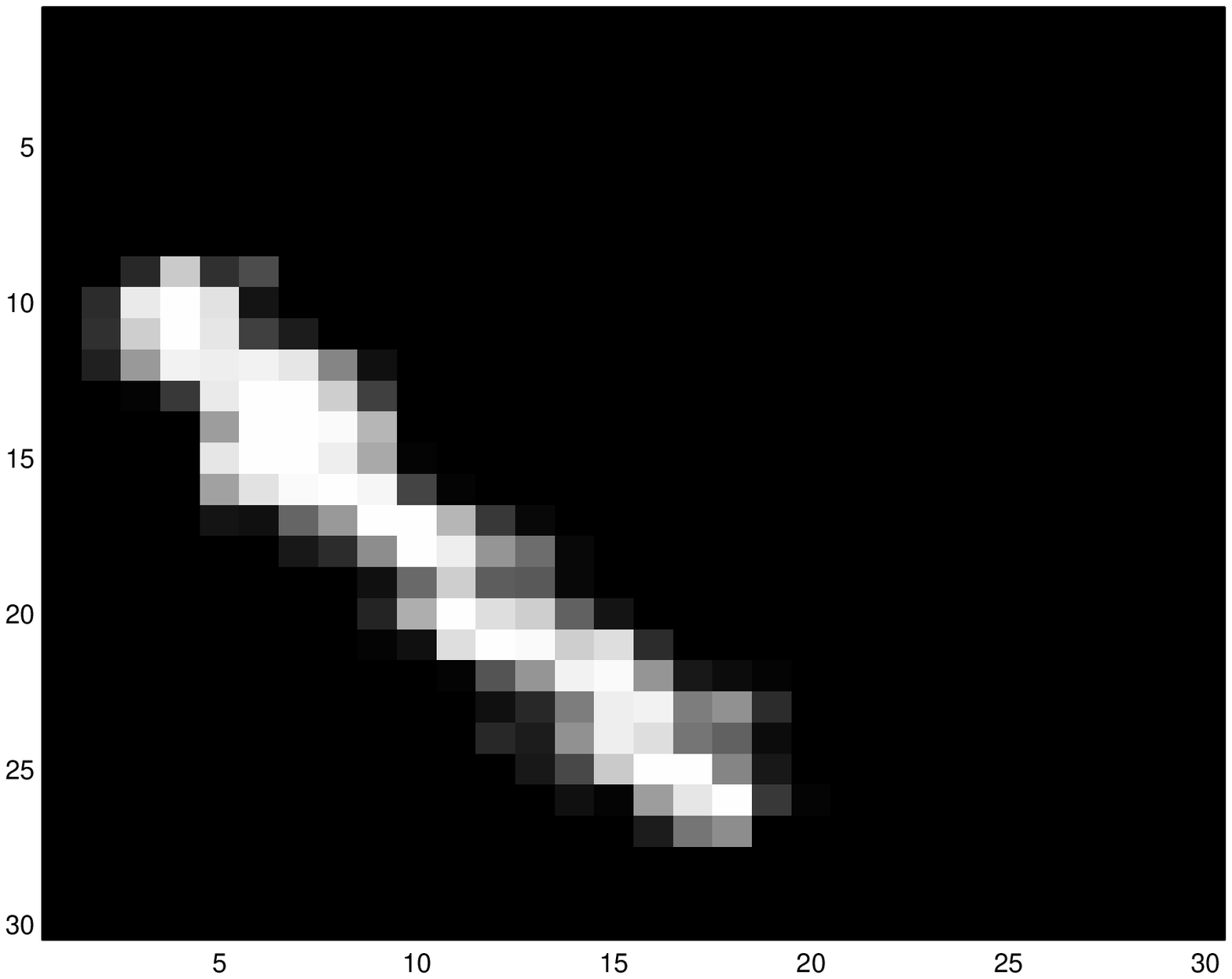}
  \includegraphics[width=.1\textwidth]{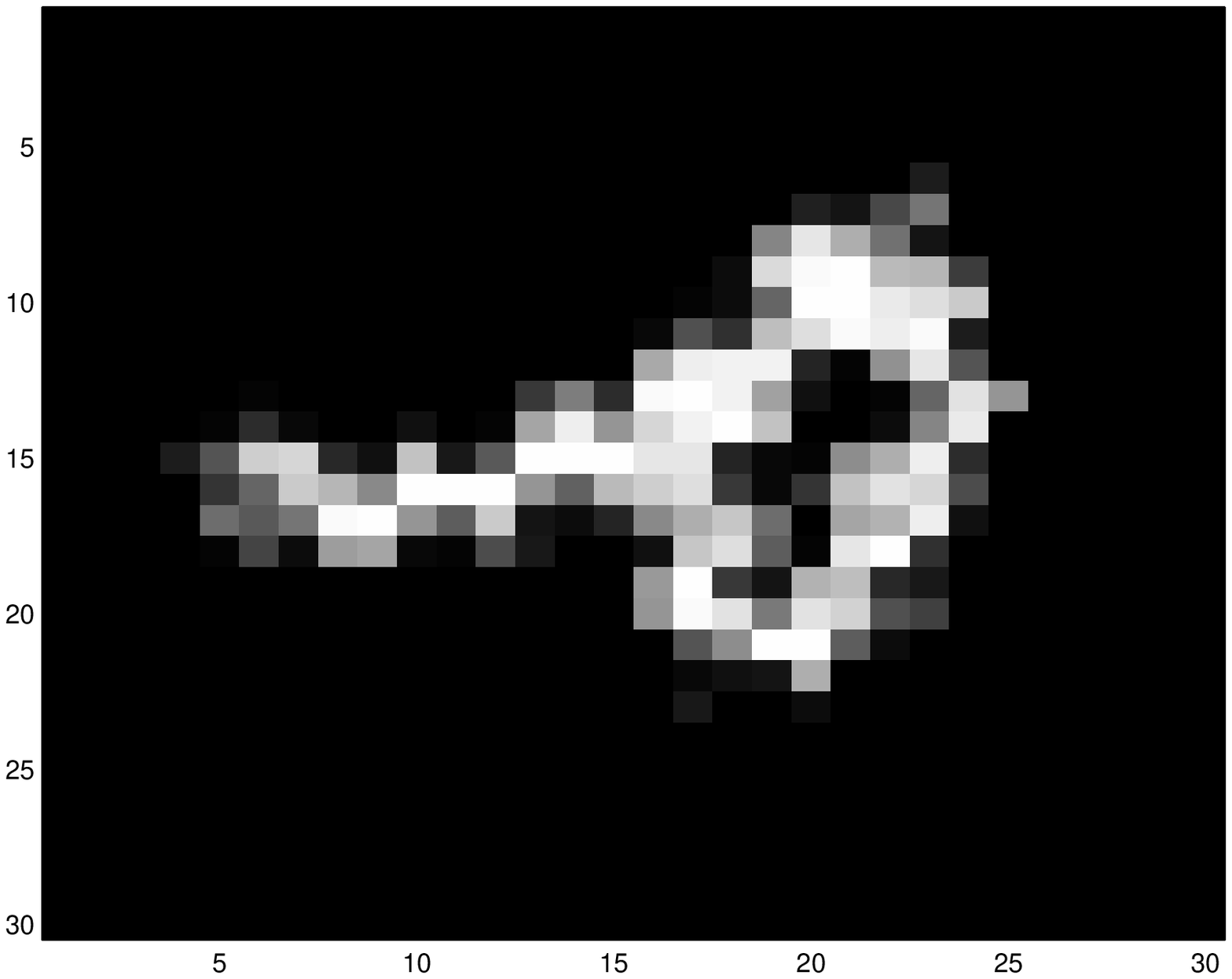}
  \includegraphics[width=.1\textwidth]{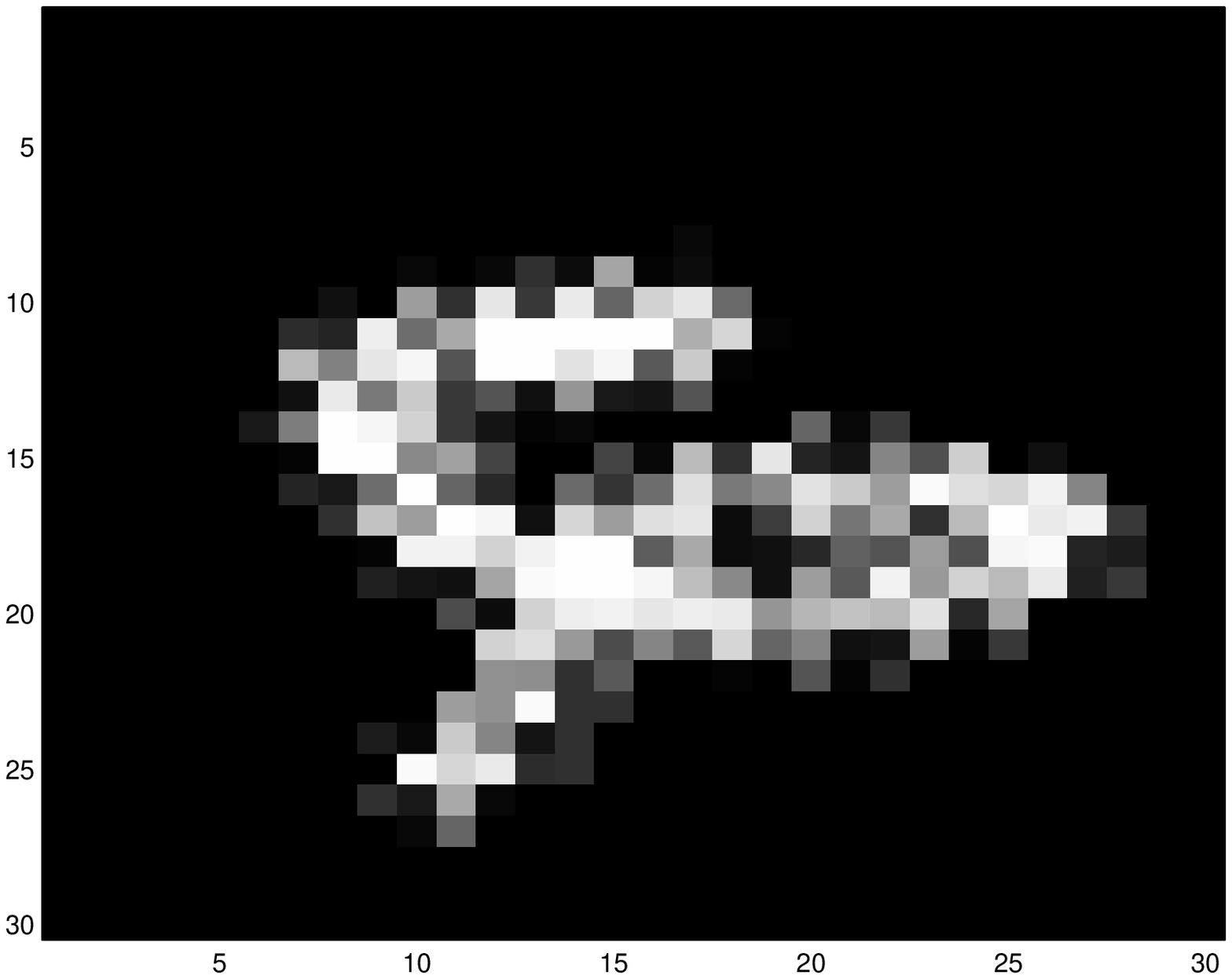}
  \includegraphics[width=.1\textwidth]{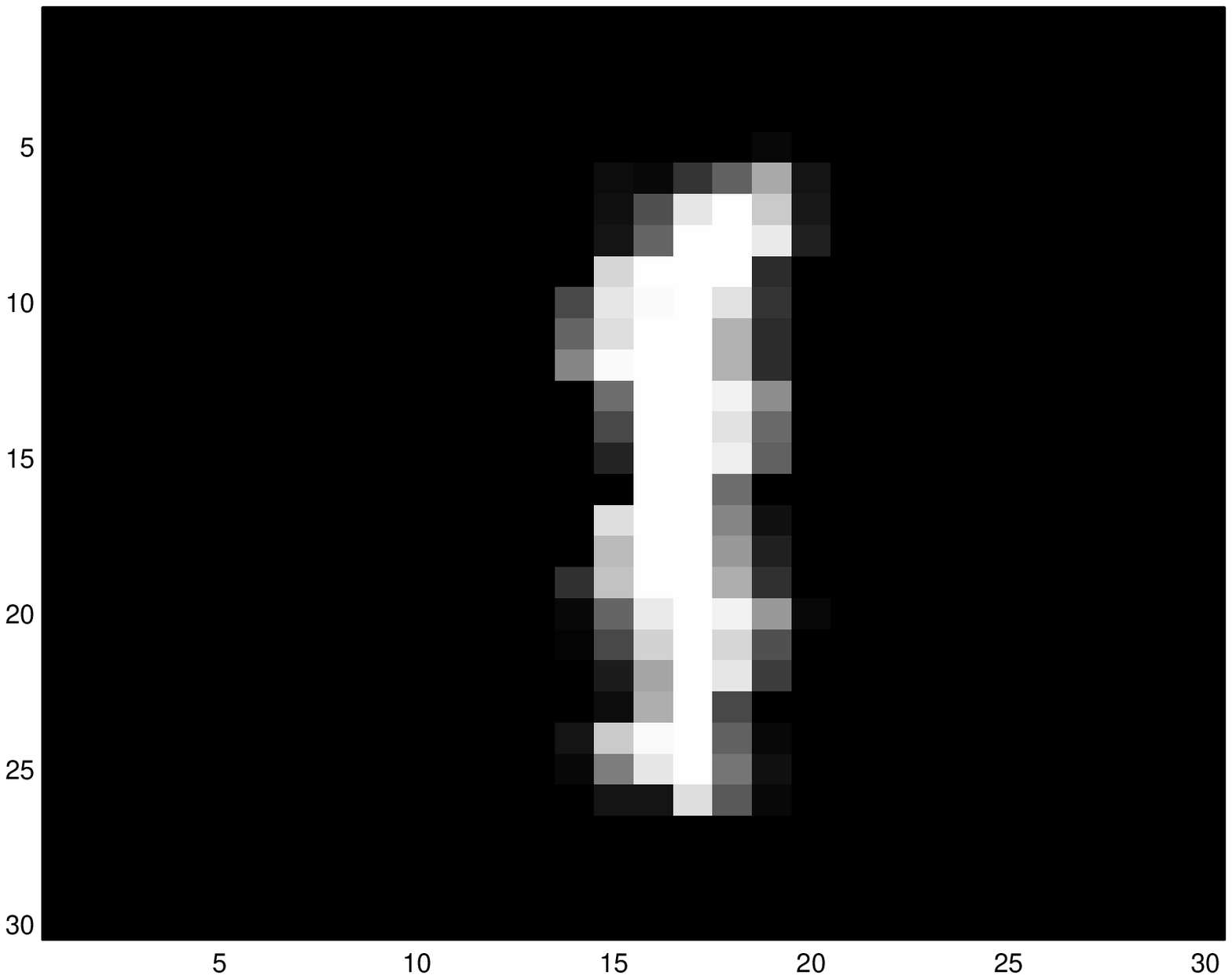}
  \includegraphics[width=.1\textwidth]{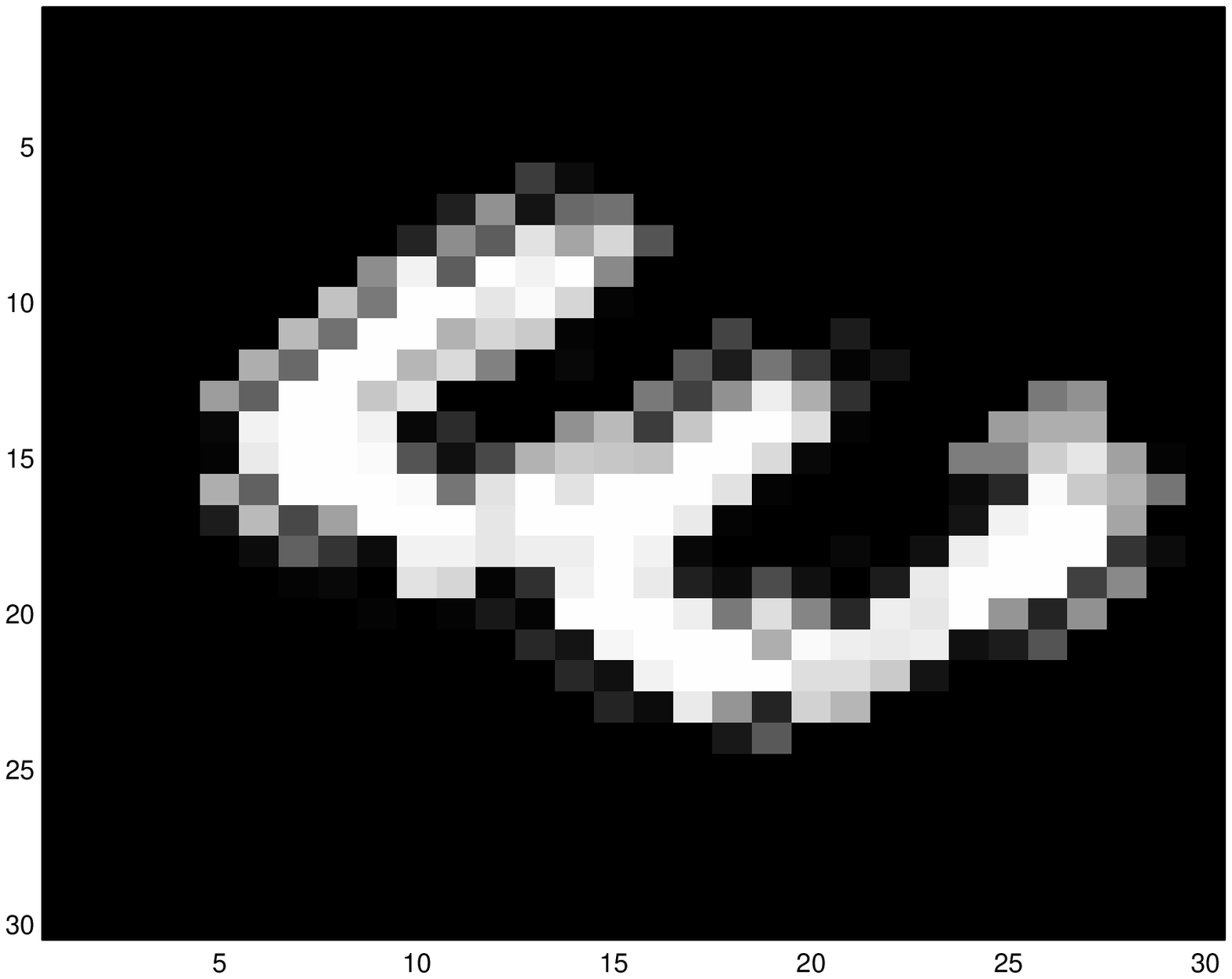}
  \caption{The first few rotated and translated NIST characters.}
  \label{fig: samples}
\end{figure*}

The bispectrum is derived by considering the 
 \m{(2l_1\<+1)(2 l_2\<+1)}-dimensional tensor product vectors \m{\V{\h f}_{l_1} \otimes \V{\h f}_{l_2}}, which  
transform according to 
\begin{equation}\label{eq: prod transform}
  \V{\h f}_{l_1} \otimes \V{\h f}_{l_2} \mapsto 
  \brbig{D^{(l_1)}(R) \otimes D^{(l_2)}(R)} \cdot 
  \brbig{\V{\h f}_{l_1} \otimes \V{\h f}_{l_2}}. 
\end{equation}
\ignore{
Instead, we now make recourse to a powerful theorem from representation theory, 
which states that for any representation \m{D} of a compact group \m{G}, there is an invertible square matrix \m{T} 
decomposing \m{D} into a sum of irreducible representations:
\begin{equation*}
  D(g)=T^{-1} \sqbBig{\bigoplus_i D^{(i)}(g)}\: T\qqquad \forall\:g\in G. 
\end{equation*}
In particular, the tensor product representation of \rf{eq: prod transform} may be reduced to such a block-diagonal form,  
as is well known from amongst other fields, quantum mechanics, where it plays a pivotal role in the theory of spin and 
angular momentum. For \m{\SO(3)} the transformation is given by 
}

The representation theory of \m{\SO(3)} is well developed, in particular, it is well known that the 
tensor product representations decompose in the form 
\begin{multline*}
  D^{(l_1)}(R) \otimes D^{(l_2)}(R) =\\
  \br{C^{l_1,l_2}}^\dag \sqbbigg{~\bigoplus_{l=\abs{l_1-l_2}}^{l_1+l_2} D^{(l)}(R)~}\; C^{l_1,l_2}.  
\end{multline*}
Here \m{{C^{l_1,l_2}}} is a \m{((2l_1\<+1)(2l_2\<+1)) \times ((2l_1\<+1)(2l_2\<+1))}-element unitary matrix, 
with rows labeled by the pair \m{\br{l,m}} and columns labeled by the pair \m{\br{m_1,m_2}}.  
The matrix elements \m{C^{l_1,l_2,l}_{m_1,m_2,m}=\sqb{C^{l_1,l_2}}_{(l,m),(m_1,m_2)}} 
are called \df{Clebsch-Gordan coefficients}, and are implemented in most computational algebra packages. 
Our notation is redundant in that it is possible to show that \m{C^{l_1,l_2,l}_{m_1,m_2,m}} vanishes 
unless \m{m_1\< +m_2\< =m}, hence we only need to worry about the coefficients 
\m{C^{l_1,l_2,l}_{m_1,m-m_1,m}}. 

Thus, under rotation \m{ C^{l_1,l_2} \brbig{\V{\h f}_{l_1} \<\otimes \V{\h f}_{l_2}}} transforms according to 
\begin{equation}\label{eq: prod transform}
  C^{l_1,l_2} \brbig {\V{\h f}_{l_1} \<\otimes \V{\h f}_{l_2}} \mapsto 
  \sqbbigg{~\bigoplus_{l=\abs{l_1-l_2}}^{l_1+l_2} D^{(l)}(R)~}\; C^{l_1,l_2}\: 
  \brbig{\V{\h f}_{l_1} \<\otimes \V{\h f}_{l_2}}.   
\end{equation}
Writing  
\m{C^{l_1,l_2}\brbig{\V{\h f}_{l_1} \<\otimes \V{\h f}_{l_2}}= \bigoplus_{l=\abs{l_1-l_2}}^{l_1+l_2} \glll{l_1}{l_2}{l}}, 
where 
\begin{equation*}
  \sqb{\,\glll{l_1}{l_2}{l}\,}_m= \sum_{m_1=-l_1}^{l_1}  
  C^{l_1,l_2,l}_{m_1,m-m_1,m} \, \flm{l_1}{m_1}\, \flm{l_2}{m-m_1}, 
\end{equation*}
\m{\glll{l_1}{l_2}{l}} transforms according to 
\begin{equation*}
  {\glll{l_1}{l_2}{l}\mapsto D^{(l)}(R) \, \glll{l_1}{l_2}{l}}. 
\end{equation*}  
By the same  argument as for the power spectrum, this gives rise to the cubic invariants 
\begin{multline}\label{eq: bispectrum}
  p_{l_1,l_2,l}= 
  \glll{l_1}{l_2}{l}^\dag \cdot \h{\V f}_{l}=\\ 
  \sum_{m=-l}^l \sum_{m_1=-l_1}^{l_1}  C^{l_1,l_2,l}_{m_1,m-m_1,m}\, \flm{l_1}{m_1}^\ast\, \flm{l_2}{m-m_1}^\ast\, \flm{l}{m}. 
\end{multline}
Up to unitary transformation, these invariants are equivalent to the non-vanishing matrix elements of 
the abstract bispectrum (as already derived in \cite{KakaralaPhD} and \cite{healy96ffts}). 
Any kernel built from the bispectrum using \rf{eq: bispectrum} as features will be invariant to 
translation and rotation.

\subsection{Computational considerations}

\begin{table*}[t]
  \tiny
  \centering
  \begin{tabular}{|l|l|l|l|l|l|l|l|l|l|l|}
\hline
&1&2&3&4&5&6&7&8&9\\
\hline
0&${0.77(0.41)}$&${6.22(2.41)}$&${5.09(1.54)}$&${5.03(1.07)}$&${2.90(1.53)}$&${4.11(2.39)}$&${2.73(1.11)}$&${4.98(1.64)}$&${5.86(2.88)}$\\
&${17.12(3.67)}$&${33.87(3.59)}$&${42.06(3.59)}$&${30.64(2.53)}$&${37.82(3.51)}$&${31.42(5.85)}$&${29.36(3.83)}$&${42.58(4.33)}$&${27.61(3.16)}$\\
\hline
1&&${0.68(0.81)}$&${0.39(0.98)}$&${3.07(1.30)}$&${0.00(0.00)}$&${1.37(0.88)}$&${1.77(1.48)}$&${2.68(2.02)}$&${1.02(1.00)}$\\
&&${30.78(2.90)}$&${29.34(4.50)}$&${34.96(3.41)}$&${30.66(2.85)}$&${34.46(4.47)}$&${38.32(4.05)}$&${24.60(2.57)}$&${34.78(3.57)}$\\
\hline
2&&&${15.89(5.79)}$&${15.82(3.22)}$&${8.06(3.60)}$&${9.64(2.00)}$&${11.11(2.29)}$&${9.26(1.63)}$&${10.55(2.95)}$\\
&&&${49.06(4.18)}$&${47.12(4.72)}$&${45.20(4.26)}$&${51.44(5.21)}$&${47.20(5.54)}$&${47.44(6.23)}$&${46.70(2.95)}$\\
\hline
3&&&&${4.81(1.68)}$&${16.42(5.69)}$&${7.54(2.75)}$&${4.00(1.13)}$&${10.70(3.79)}$&${7.66(3.01)}$\\
&&&&${44.64(3.03)}$&${49.07(4.81)}$&${49.38(5.26)}$&${44.74(4.42)}$&${50.37(4.21)}$&${47.60(5.55)}$\\
\hline
4&&&&&${6.26(1.90)}$&${10.94(4.09)}$&${14.95(2.89)}$&${6.27(3.57)}$&${16.95(1.84)}$\\
&&&&&${40.08(6.67)}$&${50.11(5.26)}$&${45.30(3.30)}$&${46.26(2.63)}$&${49.82(4.68)}$\\
\hline
5&&&&&&${14.63(2.42)}$&${5.31(2.27)}$&${6.62(2.72)}$&${6.84(2.23)}$\\
&&&&&&${50.00(4.02)}$&${41.70(4.09)}$&${44.63(3.31)}$&${46.01(4.37)}$\\
\hline
6&&&&&&&${7.68(4.05)}$&${9.00(2.93)}$&${20.15(3.62)}$\\
&&&&&&&${48.19(4.10)}$&${46.13(5.82)}$&${53.75(2.69)}$\\
\hline
7&&&&&&&&${3.50(2.28)}$&${8.06(3.49)}$\\
&&&&&&&&${41.16(5.18)}$&${53.21(5.01)}$\\
\hline
8&&&&&&&&&${9.43(2.14)}$\\
&&&&&&&&&${45.13(2.87)}$\\
\hline
  \end{tabular}
  \label{tbl: results lin}
  \caption{Classification error in percent for each pair of digits for the linear kernels. 
  The performance of the bispectrum-based classifier is shown on top, and the baseline on bottom;  
  standard errors are in parentheses.
}
\end{table*}

The algorithmic implementation of \rf{eq: bispectrum} is  
\vspace{-15pt}
\begin{multline*}
  p_{l_1,l_2,l}= 
  \sum_{m=-l}^l \flm{l}{m} \times \\ 
  \times \sum_{m_1=\max(-l_1,m-l_2)}^{\min(l_1,m+l_2)}  C^{l_1,l_2,l}_{m_1,m-m_1,m}\, \flm{l_1}{m_1}^\ast\, {\flm{l_2}{m-m_1}^\ast}, 
\end{multline*}
which gives \m{O(L^3)} invariant features to build the kernel from. 
The features can be precomputed as a data processing step before any learning actually takes place. 
Typically, \m{L} will scale linearly with the linear dimension \m{w} of the input image in pixels, 
so the bispectrum inflates the data at a rate of \m{u^{3/2}}, where \m{u} is the original storage size of 
a single image. 

Projecting onto the sphere is a linear map and its coefficients can be precomputed, so the cost of that operations scales 
with \m{w^2 L^2\propto u^2}. 
Finally, computing the bispectrum itself scales with \m{L^5\propto u^{5/2}}. 
On the desktop PC used to prepare the data for the experiments, processing each \m{30\times 30} pixel 
image took approximately 100ms for \m{L=15}.

\ignore{ 

These invariants form the non-vanishing components of what is known as the  \df{bispectrum} of \m{f}. 
The bispectrum is widely known in the context of signals on \m{[0,2\pi)} or the discrete 
set \m{\zton{n\<-1}}, but its generalization to homogeneous spaces of compact groups has attracted less attention.   
The seminal work 
in this field 
is by Kakarala \cite{KakaralaPhD},  
who presents the case of functions on \m{S_2} as an example, 
but discusses it in a slightly different form from our exposition above.    
In the context of fast Fourier transforms on the sphere, the bispectrum is also mentioned in 
\cite{healy96ffts} and \cite{kostelec-computational}. 
The most significant and surprising result from \cite{KakaralaPhD}, however, is that up to the action of the underlying group 
(in our case, rotations),   
the invariants \m{\br{p_{l_1,l_2,l}}_{l_1,l_2=0,1,2,\ldots;~ l=\abs{l_1-l_2},\ldots,l_1+l_2}} 
are sufficient to reconstruct the original function \m{f}. 
As a direct corollary, for band-limited functions  
just the invariants \m{\br{p_{l_1,l_2,l}}_{0\leq l_1,l_2 \leq L}} are sufficient.

To construct an \m{\SO(3)}-invriant kernel between two functions \m{f_1} and \m{f_2} on \m{S_2}, 
we use the real and imaginary parts of the 
components \m{\brN{p^{(1)}_{l_1,l_2,l}}} and \m{\brN{p^{(2)}_{l_1,l_2,l}}} of their respective bispectra 
as features and use a conventional kernel on the resulting Euclidean space. 
In our experiments we simply used the dot-product kernel 
\begin{equation*}
  k(f_1,f_2)=\sum_{l_1=0}^{L}~\sum_{l_1=0}^{L}~\sum_{l=\abs{l_1-l_2}}^{l_1+l_2} 
  \br{\RE\brN{p^{(1)}_{l_1,l_2,l}} \,\cdot \:\RE\brN{p^{(2)}_{l_1,l_2,l}} 
    + {\IM\brN{p^{(1)}_{l_1,l_2,l}} \,\cdot \:\IM\brN{p^{(2)}_{l_1,l_2,l}} } }. 
\end{equation*}

}

\begin{table*}[]
  \tiny
  \centering
  \begin{tabular}{|l|l|l|l|l|l|l|l|l|l|l|}
\hline
&1&2&3&4&5&6&7&8&9\\
\hline
0&${0.80(0.42)}$&${5.06(1.52)}$&${4.78(1.08)}$&${3.35(1.69)}$&${3.90(2.25)}$&${3.07(1.77)}$&${4.48(1.39)}$&${3.74(2.23)}$&${6.34(2.57)}$\\
&${12.50(3.60)}$&${26.30(4.32)}$&${33.72(4.58)}$&${32.45(12.63)}$&${29.52(3.99)}$&${23.51(4.93)}$&${24.96(3.73)}$&${29.99(4.20)}$&${19.16(2.65)}$\\
\hline
1&&${0.99(0.48)}$&${0.00(0.00)}$&${2.48(0.97)}$&${0.21(0.45)}$&${1.35(0.43)}$&${1.22(1.09)}$&${0.52(0.55)}$&${3.05(0.88)}$\\
&&${27.29(4.00)}$&${22.61(8.82)}$&${33.98(9.44)}$&${30.86(9.99)}$&${28.52(9.47)}$&${32.12(6.34)}$&${20.16(2.93)}$&${28.01(4.56)}$\\
\hline
2&&&${14.68(4.60)}$&${13.20(2.56)}$&${8.83(4.22)}$&${8.89(3.09)}$&${12.73(3.39)}$&${12.14(2.27)}$&${10.34(2.51)}$\\
&&&${47.75(3.46)}$&${45.26(5.11)}$&${50.09(4.78)}$&${45.63(5.49)}$&${43.84(4.38)}$&${44.02(3.14)}$&${45.95(4.84)}$\\
\hline
3&&&&${5.12(2.35)}$&${16.88(2.73)}$&${6.98(3.46)}$&${3.50(1.48)}$&${10.21(3.89)}$&${5.08(1.50)}$\\
&&&&${43.07(9.05)}$&${52.53(3.39)}$&${45.86(5.27)}$&${41.90(4.09)}$&${46.00(4.97)}$&${44.87(3.91)}$\\
\hline
4&&&&&${5.75(1.22)}$&${10.67(1.47)}$&${13.92(2.63)}$&${6.45(2.26)}$&${12.09(2.47)}$\\
&&&&&${39.21(4.29)}$&${46.82(5.32)}$&${46.73(6.47)}$&${42.29(4.44)}$&${52.73(3.65)}$\\
\hline
5&&&&&&${16.56(1.66)}$&${6.26(1.54)}$&${6.23(3.05)}$&${7.07(2.93)}$\\
&&&&&&${47.04(4.21)}$&${46.39(3.41)}$&${41.63(3.29)}$&${43.23(2.46)}$\\
\hline
6&&&&&&&${9.30(3.33)}$&${6.16(2.30)}$&${21.37(3.81)}$\\
&&&&&&&${40.43(5.16)}$&${41.19(4.47)}$&${50.73(4.31)}$\\
\hline
7&&&&&&&&${4.68(2.30)}$&${8.81(2.81)}$\\
&&&&&&&&${37.33(2.21)}$&${46.22(4.13)}$\\
\hline
8&&&&&&&&&${10.06(2.04)}$\\
&&&&&&&&&${44.06(3.93)}$\\
\hline
  \end{tabular}
  \label{tbl: results rbf}
  \caption{Classification error in percent for each pair of digits for the Gaussian RBF kernels. 
  The performance of the bispectrum-based classifier is shown on top, and the baseline on bottom, 
  standard errors are in parentheses.
}
\end{table*}

\section{Experiments}

We conducted experiments on randomly translated and rotated versions of hand-written digits from 
the well known NIST dataset \cite{NISTLeCun}. 
The original images are size \m{28\times 28}, but most of them only occupy a fraction of the image patch. 
The characters are rotated by a random angle between \m{0} and \m{2\pi}, clipped, and embedded 
at a random position in a \m{30\times 30} patch (fig. \ref{fig: samples}). 

We trained 2-class SVMs for all possible pairs of digits. 
As a baseline we used SVMs with linear and Gaussian RBF kernels on the original \m{900}-dimensional 
pixel intensity vector. 
We compared this to similar linear and Gaussian RBF SVMs ran on the bispectrum features. 
We used \m{L=15}, which is a relatively low resolution for images of this size. The magnification parameter was set 
to \m{a=2}. 

Our experimental procedure consisted of using cross-validation to set the regularization parameter 
\m{C} and the kernel width \m{\sigma} independently for each 
each learning task: digit \m{d_1} vs. digit \m{d_2}. 
We used \m{10}-fold cross validation to set the parameters for the linear kernels, but to save time only 
\m{3}-fold cross validation for the Gaussian kernels.
Testing and training was conducted on the relevant digits from the second one thousand images in the NIST dataset. 
The results we report are averages and standard deviations of error for 10 random even splits of this data. Since there 
are on average \m{100} digits of each type amongst the \m{1000} images in the data, our average training set and test set consisted of 
just 50 digits of each class. Given that the images also suffered random translations and rotations this is 
an artificially difficult learning problem.

The results are shown in table \ref{tbl: results lin} for the linear kernel and in table \ref{tbl: results rbf} for the 
RBF kernel. The two sets of results are very similar. 
In both cases the bispectrum features far outperform the baseline bitmap representation. 
Indeed, it seems that in many cases 
the baseline cannot do better than what is essentially random guessing. 
In contrast, the bispectrum can effectively discriminate even in the hard cases such as 8 vs. 9 and 
reaches almost 100\% accuracy on the easy cases such as \m{0} vs. 1. 
Surprisingly, to some extent 
the bispectrum can even discriminate between 6 and 9, which in some fonts are exact rotated versions of each other. 
However, in handwriting, 9's often have a straight leg and/or a protrusion at the top where right handed scribes reverse the 
direction of the pen. 

The results make it clear that the bispectrum features are able to capture position and orientation invariant 
characteristics of handwritten figures. 
We did not compare our algorithm against other image kernels due to time constraints. 
However, short of a handwriting-specific algorithm which extracts exlicit landmarks 
we do not expect other methods to yield a comparable degree of position and rotation invariance.

\section{Conclusions}

We presented an application of the theory of bispectra on non-commutative groups to constructing 
a novel system of translationally and rotationally invariant features for images. 
The method hinges on a  projection from the plane to the sphere, reducing the problem 
of invariance to the action of the non-compact Euclidean group to that of the compact and 
computationally tractable three dimensional rotations group. 

Our method may be used as a pre-processing step for learning algorithms, 
in particular, kernel-based discriminative algorithms. 
Computational requirements scale with \m{u^{5/2}} and memory requirements with \m{u^{3/2}} where 
\m{u} is the size of the original image (in pixels). 

Experimental results on an optical character recognition problem indicate that the method is surprisingly 
powerful ``out of the box''. Time constraints prevented us from conducting more extensive experiments on 
larger images (entire scenes), multicolor images, etc., but we expect our algorithm to remain viable 
over a range of tasks.

Finally, we believe that the general concept of bispectra ought to be of interest to the machine learning community as 
it moves towards addressing learning tasks on more and more intricately structured data. 
This motivated the general discussion of the bispectrum concept in the first half of this paper. 


\section{Acknowledgments}

I am indebted to G{\'a}bor Cs{\'a}nyi for drawing my attention to the bispectrum. 
I would also like to thank Ramakrishna Kakarala for providing me with a copy of his doctoral thesis, and 
Ron Dror, Tony Jebara, Albert Bart{\'o}k-Partay and Bal{\'a}zs Szendr{\H o}i for discussions. 
This work was supported in part by National Science
Foundation grants IIS-0347499, CCR-0312690 and IIS-0093302.

\bibliographystyle{plain}
\bibliography{invImagesAISTATS07}

\end{document}